\documentclass{article}
\usepackage{iclr2027_conference,times}
	
\usepackage{amsmath,amsfonts,bm}

\def\eqref#1{equation~\ref{#1}}

\def\1{\bm{1}}

\DeclareMathAlphabet{\mathsfit}{\encodingdefault}{\sfdefault}{m}{sl}
\SetMathAlphabet{\mathsfit}{bold}{\encodingdefault}{\sfdefault}{bx}{n}

\usepackage{hyperref}
\usepackage{url}
\usepackage{graphicx}
\usepackage{booktabs}
\usepackage{multirow}
\usepackage{subcaption}
\usepackage{xcolor}

\usepackage{enumitem}

\graphicspath{{figures/}}

\newcommand{\method}{Latent Harness}
\newcommand{\ours}{\method}

\title{\method{}: Compiling Agent Behaviors into Post-Block Activation Shifts}
\author{
	Ran Li\textsuperscript{1,§} \quad
	Shimin Di\textsuperscript{3} \quad
	Zeyu Ren\textsuperscript{3} \quad
	Ling Yue\textsuperscript{4} \quad
	Jiachuan Wang\textsuperscript{5} \quad
	Lei Chen\textsuperscript{1,2,‡} \\
	\textsuperscript{1}Hong Kong University of Science and Technology  \quad 
	\textsuperscript{3}Southeast University  \\
	\textsuperscript{2}Hong Kong University of Science and Technology(Guangzhou)  \\
	\textsuperscript{4}Rensselaer Polytechnic Institute \quad 
	\textsuperscript{5}University of Tsukuba \\
	\textsuperscript{§}\texttt{rlibb@connect.ust.hk} \quad \textsuperscript{‡}\texttt{leichen@cse.ust.hk}
}
\iclrfinalcopy

\begin{document}

\maketitle
\makeatletter
\def\@oddhead{\small Preprint. Under review.\hfil}
\def\@evenhead{\small Preprint. Under review.\hfil}
\makeatother
\begin{abstract}
Harness engineering and self-evolving agents improve language-model agents quickly, yet their gains live outside the model. The harness is re-sent and re-paid on every call, and the standard route to internalization is a weight update that is costly to iterate and fixed once merged. We propose compiling the harness instead. \ours{} distills an external condition into a rank-8 activation shift. The shift is applied at the output of every decoder block of a frozen backbone, at 0.03--0.05\% of parameters. The condition is either an explicit process-level behavior instruction or the behavioral contract of an agentic harness. At inference the compiled bundle carries no harness text and adds no tokens. It exposes three runtime controls for injection strength, injection phase, and shift-space mixing. Compiled behaviors recover the source-harness gain in 8 of 10 preregistered settings. Where the explicit source pays the heaviest per-call token tax, the compiled form exceeds it. The same bundles transfer zero-shot across tasks (5 of 7 pairs positive) and compose in shift space, while stacked text instructions interfere destructively. Used as a search space, they lift budget-matched harness search by 1.45$\times$. On three agentic benchmarks the operator aligns policy to the harness contract. Under matched data, capacity, and effective strength it outperforms same-capacity LoRA, which needs 11.1$\times$ the parameters to reach the same band, and it regresses nothing where the base policy is already strong. Across 1B-to-8B models the method skeleton is invariant while gain locations move with the model-task pair, which delimits where compiled harnesses should be deployed. Our code is available at \url{https://github.com/rlisml/latentharness}.
\end{abstract}

\section{Introduction}
\label{sec:intro}

The performance of a language-model agent is increasingly set by its harness, meaning the instructions, tool policies, memory, and verification scaffolds wrapped around model calls. Recent work searches over program spaces \citep{hu2025automated,zhang2025aflow} and over modular design spaces \citep{shang2025agentsquare,zhang2025multi}, and some of it constrains the search to safety \citep{zheng2025mermaidflow}. A second line makes harnesses improve themselves, either by synthesizing code harnesses from failures \citep{lou2026autoharness}, by editing executable runtime harnesses \citep{shao2026harness}, or by optimizing harness and model jointly \citep{lee2026meta}. Surveys document how broad this engineering surface has become \citep{meng2026agent}. Where the resulting gains end up is almost always one of two places, in text that surrounds the model or in the model's weights.

Both carry structural costs. A textual condition is a recurring expense. Even a 49-token process instruction is re-sent on every call, the behavior exists only while the text is present, and serving several harnesses multiplies prompt length. Weight updates internalize at a different price. Training-style policy alignment such as SFT, RLOO, and GRPO \citep{shao2024deepseekmath}, co-evolution frameworks that update policy weights alongside the harness \citep{chen2026co,chen2026harnessforgejointharnesspolicy,hebbar2026sia}, and self-evolving agents that write acquired skills into fine-tuned or RL-trained weights \citep{huang2026r,xia2026skillrl,zhang2026memskill} all modify millions of parameters per behavior. They also bind one behavior to one set of edited weights and fix the intervention strength at merge time.

\begin{figure}[t]
\centering
\includegraphics[width=0.9\linewidth]{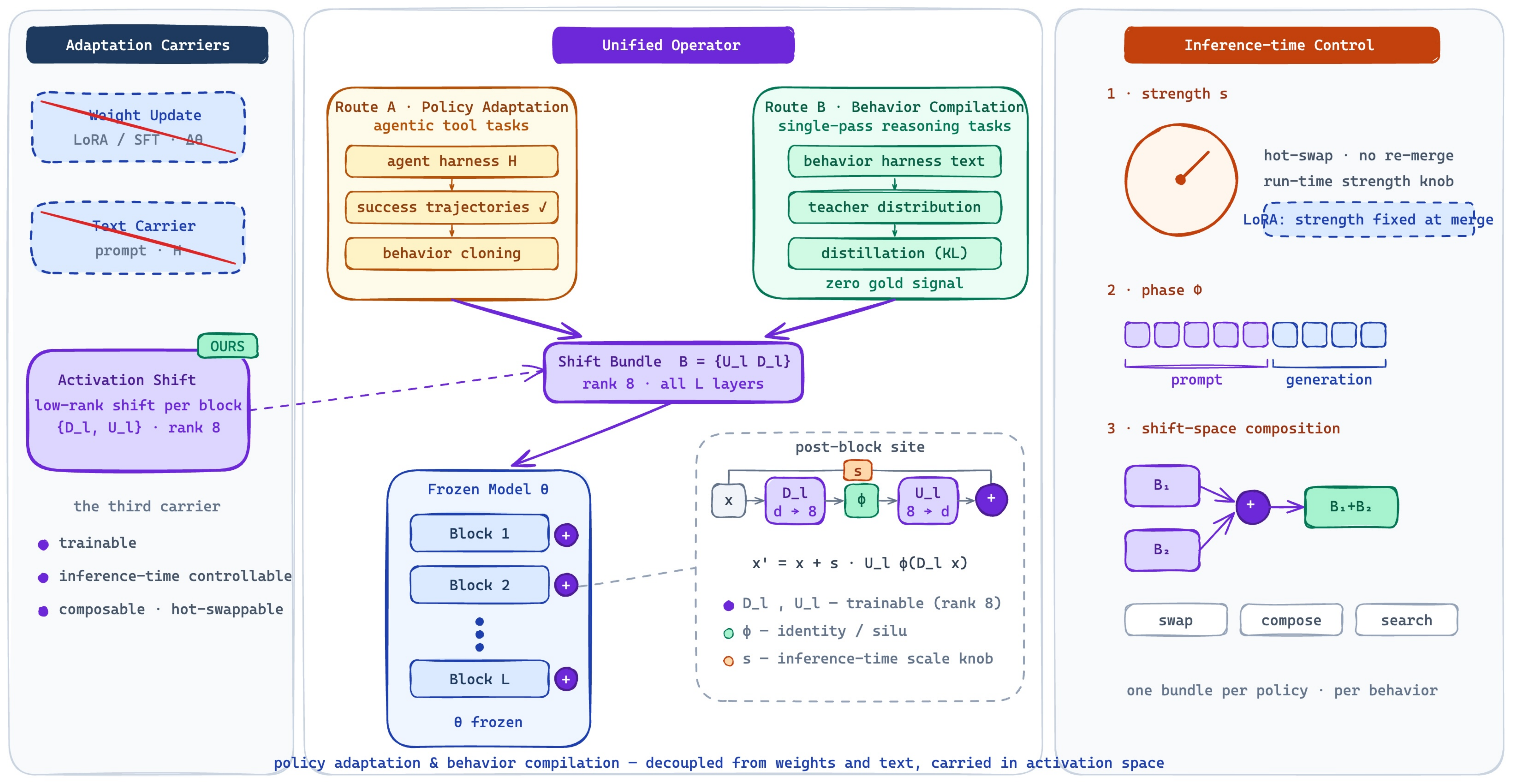}
\caption{\textbf{\ours{} compiles external conditions into post-block activation shifts.} Teacher traces or strictly filtered success trajectories train a rank-8 shift at every block output of a frozen backbone. The bundle injects at run time with strength, phase, and mixing controls.}
\label{fig:overview}
\end{figure}

We propose a third destination and compile the harness. At compile time the frozen model produces its own supervision under the external condition. The supervision is either teacher traces generated with an explicit behavior instruction prepended, or decision pairs harvested from trajectories that succeeded under an agentic harness. A rank-8 low-rank shift is then trained at the output of every decoder block and exported as a per-layer bundle. At run time the bundle is injected into the residual stream after each block. The model then behaves as if the condition were present, without the condition text, at zero additional tokens, with every backbone parameter bit-identical. We call the compiled artifact a \emph{latent harness}. Figure~\ref{fig:overview} summarizes the pipeline.

\noindent Compilation preserves behavior more faithfully than we expected. Across preregistered settings the compiled shift recovers most of the explicit-harness gain. In the settings where the text harness pays the heaviest per-call token tax, the compiled form exceeds it, because the internalized behavior pays that tax once, at compile time. The compiled object is the behavior itself and does not reduce to generic task skill. Preregistered behavioral markers move with the scores, accuracy-only adapters dissociate from behavior, shuffled pairings fail to reproduce it, and a harmful harness compiles into a harmful latent with the same fidelity. Compiled bundles are also reusable. They transfer zero-shot across tasks, they compose in shift space while stacked text instructions interfere destructively, and they lift budget-matched harness search by 1.45$\times$. The same operator aligns policy on three agentic benchmarks. Under matched data, capacity, and effective strength it outperforms same-capacity LoRA, and on a benchmark where the base policy is already strong it regresses nothing. Our contributions are as follows.
\begin{itemize}[leftmargin=*,labelindent=0pt,labelsep=0.5em,itemsep=0pt,topsep=1pt,parsep=0pt]
\item \textbf{A compilation abstraction for harness gains.} One operator, a post-block rank-8 shift on a frozen backbone at 0.03--0.05\% of parameters, compiles two kinds of external conditions into an artifact that needs no condition text and no tokens at inference. The artifact exposes runtime controls for strength, phase, and mixing.
\item \textbf{Evidence that compilation preserves the behavior.} 8 of 10 preregistered settings recover at least half of the source gain. A three-chain protocol over scores, preregistered markers, and controls attributes the recovery, including symmetric replication of harmful signatures. Where the text source pays the heaviest token tax, the compiled form exceeds it.
\item \textbf{Evidence that compiled bundles are reusable.} Frozen bundles transfer zero-shot across tasks (5 of 7 pairs positive, 59--74\% out-of-distribution recovery). They compose exactly in shift space, where text stacking is destructive, and they deliver a 1.45$\times$ search gain at a matched evaluation budget.
\item \textbf{Policy compilation on agent benchmarks.} At 1.012$\times$ parameters and matched data and strength, the shift beats same-capacity LoRA on ToolHop (60.00\% vs.\ 54.74\%, paired net $+5$). Full-capacity LoRA needs 11.1$\times$ (4B) and 3.09$\times$ (8B) parameters to reach the same band. API-Bank shows zero regression. The skeleton is invariant from 1B to 8B while gains move with the model-task pair.
\end{itemize}

\section{Compiling behaviors into post-block shifts}
\label{sec:method}

\subsection{The post-block injection site}
\label{sec:operator}

Let the frozen policy be an autoregressive transformer $\pi_\theta$ with $L$ decoder blocks. In residual-stream notation, block $l$ computes $x^{l+1} = x^{l} + f_l(x^{l})$. We inject at the \emph{post-block} site, the point where everything block $l$ computed meets everything blocks $l{+}1,\dots,L$ will read. For every layer we train a down projection $D_l \in \mathbb{R}^{r\times d}$ and an up projection $U_l \in \mathbb{R}^{d\times r}$ with rank $r{=}8$, and replace the block output by
\begin{equation}
\tilde{x}^{l+1}_i \;=\; x^{l+1}_i \;+\; s\,\alpha\, U_l\big(\varphi(D_l\, x^{l+1}_i)\big)\,\mathbf{1}[i\in\Phi],
\label{eq:inject}
\end{equation}
where $s$ is a global runtime strength, $\alpha$ is a fixed training scale, $\varphi$ is a bottleneck nonlinearity, and the phase set $\Phi$ selects which token positions receive the shift. Trainable parameters total $2rdL$, between 0.029\% and 0.042\% of the backbone across the five models we study (Table~\ref{tab:setup} in the appendix).

Two properties make this site the right compilation target. A perturbation of the residual trunk is read by every subsequent layer, so one parameter influences far more computation than a perturbation inside a single linear map, which propagates only within its layer. The site is also the unique junction at which any first-order correction to a block's behavior can be expressed, which explains why two very different compile sources succeed under the same operator. \citet{adila2026weightupdatesactivationshifts} show that weight updates can be viewed as activation shifts. We compile conditions that never required a weight update into the same substrate. The operator is first-order kin to LoRA \citep{hu2021lora}, since both inject rank-at-most-$r$ perturbations, which is what makes capacity-aligned comparison possible (Section~\ref{sec:policy}). The bottleneck nonlinearity is $\varphi=\mathrm{SiLU}$ with zero-initialized $U_l$ for instruction compilation and the identity for policy compilation. Both choices work, indicating low sensitivity to this detail.

\subsection{Compile source A: explicit behavior instructions}
\label{sec:instruction}

The first source is a registry of seven process-level behavior instructions. Each is a short prompt prefix of at most 60 tokens, single-pass and tool-free, and each targets a diagnosed error mode. The seven cover independent verification, numbered decomposition, write-the-expression-first evaluation, concise answering, option elimination, evidence quoting, and test-interface consultation. Every entry is registered with a frozen hash and carries a predicted behavioral marker, which is a deterministic parse or regex count on the raw output such as decomposition-step count or interface-alignment rate. An entry enters the study only if screening shows a significant explicit gain (Appendix~\ref{app:screening}).

The teacher is the frozen model with the instruction prepended. It produces greedy traces that are cached in full and filtered by correctness. The student is the same frozen model plus the shift of Eq.~\ref{eq:inject}, trained on prompts that are \emph{byte-identical} to the unconditioned base prompts, so the instruction text never appears in any student input (asserted per batch). The distillation loss is token-level cross-entropy on filtered teacher traces for generation tasks and an option-level KL on cached teacher answer distributions for multiple-choice tasks. No gold labels enter this configuration. Two capacity-matched controls share the budget, a gold-label task adapter and a joint objective (Appendix~\ref{app:protocol}).

\subsection{Compile source B: harness-conditioned strategies}
\label{sec:policysrc}

The second source is the behavioral contract of an agentic harness. The frozen base policy rolls out on the training pool under the official harness pairing. Each trajectory decomposes into per-step decision pairs, whose prompt is the full interaction history rendered so far and whose completion is the action taken. The \emph{strict tier} keeps pairs only from trajectories whose final answer is correct. The choice of target distribution for behavioral cloning motivates this filter \citep{torabi2018behavioral}. Failed trajectories are draws from the base policy itself, and training on them pulls the target back toward the base policy. We observed this effect directly. A balanced tier costs 3.2 accuracy points and inflates action counts by 51\%, while the strict tier gains 3.2 points with 3\% inflation. Training applies prefix-masked cross-entropy to action tokens with hyperparameters identical to the LoRA comparison configuration (Appendix~\ref{app:protocol}).

\subsection{Runtime interface}
\label{sec:runtime}

Compilation exports a bundle $\{U_l D_l\}_{l=1}^{L}$ in fp32 per layer. Load-back is bit-exact and the engine's weights are untouched. Three controls operate at inference without retraining. The global strength $s$ in Eq.~\ref{eq:inject} is hot-swappable within a serving session. The phase $\Phi\in\{\text{prompt},\text{generation},\text{both}\}$ selects which positions receive the shift. Shift-space mixing combines two compiled bundles as $\alpha Z_a + \beta Z_b$, realized by a rank-16 merged bundle with tensor-level error at most $10^{-5}$. Mixing must be defined on induced shifts rather than factor sums, because the bottleneck nonlinearity makes factor-space addition the wrong algebra (Section~\ref{sec:reusability} demonstrates the consequences). Two checks keep the measurements valid. If metrics are bit-identical across two strengths the injection is silently off, which we check per configuration. A bounded parse protocol, independent of configuration, repairs or rejects malformed tool-call JSON identically everywhere.

\section{Experiments}
\label{sec:exp}

We organize the evaluation around four questions. Does policy compilation work on agentic benchmarks, and at what parameter cost (Section~\ref{sec:policy})? Does instruction compilation preserve the explicit harness's gain (Section~\ref{sec:behavior})? What does the bundle carry, and how reusable is it (Sections~\ref{sec:what} and~\ref{sec:reusability})? Where does the intervention act, and when does it fail (Section~\ref{sec:mechanism})?

\subsection{Setup}
\label{sec:setup}

Instruction compilation uses Llama-3.2-1B-Instruct (M1), Qwen3-4B \citep{yang2025qwen3} (M2), Llama-3.1-8B-Instruct (M3), and an independent replication track on Llama-3.2-1B (base). The evaluation tasks are BoolQ \citep{clark2019boolq}, ARC-Challenge \citep{clark2018think}, GSM8K \citep{cobbe2021training}, and MBPP under sandboxed pass@1, with SVAMP and ARC-Easy as out-of-distribution targets. Policy compilation uses Qwen3-4B and Qwen3-8B on ToolHop \citep{ye2025toolhop} ($n{=}95$), SearchQA over HotpotQA \citep{yang2018hotpotqa} and 2WikiMultihopQA \citep{ho2020constructing} ($n{=}50$), and API-Bank \citep{li2023api} ($n{=}34$), all under official round-3 harness pairings with evaluation pools disjoint from training. Statistics follow one discipline. Criteria are preregistered, selection uses validation only, and the test set is touched once. We report paired bootstrap 95\% CIs ($B{=}10{,}000$) over 3 training seeds for instruction settings, and at least 2 independent full evaluations per adopted policy configuration with per-item paired net differences. Appendix~\ref{app:protocol} gives the full protocol.

\begin{table}[t]
\centering
\small
\setlength{\tabcolsep}{5.2pt}
\caption{Main results on the HarnessForge benchmark suite, following its
baseline suite. Search-style rows search over agent harnesses, training-style
rows update policy weights, and HarnessForge is the full co-evolution framework.
\textbf{Ours} is evaluated under our controlled single-round alignment
protocol, which uses the official round-3 harness pairing and evaluation pools
disjoint from training, and which is protocol-distinct from the co-evolution rows
above. Params (\%) is the trainable fraction of backbone parameters
(LoRA for training-style rows and HarnessForge, while search-style rows update no
weights). Best in \textbf{bold}, second best \underline{underlined}, per
backbone.}
\label{tab:policy-main}
\begin{tabular}{llcrrrrrrr}
\toprule
 & & & \multicolumn{2}{c}{ToolHop} & \multicolumn{3}{c}{SearchQA} & \multicolumn{2}{c}{API-Bank} \\
\cmidrule(lr){4-5}\cmidrule(lr){6-8}\cmidrule(lr){9-10}
Model & Method & Params (\%) & Ans. & Path & Hotpot & 2Wiki & Overall & Succ. & API. \\
\midrule
\multirow{10}{*}{Qwen3-4B}
 & ADAS            & 0 & 40.00 & 44.99 & 28.67 & 28.00 & 28.33 & 57.31 & 60.99 \\
 & AgentSquare     & 0 & 29.23 & 39.46 & 29.33 & 30.00 & 29.67 & 47.82 & 46.10 \\
 & AFlow           & 0 & 31.28 & 43.81 & 30.67 & 29.33 & 30.00 & 45.72 & 41.13 \\
 & MaAS            & 0 & 42.05 & 51.80 & 30.67 & 30.00 & 30.33 & 62.74 & 63.83 \\
 & MermaidFlow     & 0 & 44.62 & 58.74 & 31.33 & 28.00 & 29.67 & 64.33 & 60.28 \\
\cmidrule(lr){2-10}
 & SFT             & 0.41 & 45.13 & 63.90 & 36.67 & 38.67 & 37.67 & 75.10 & 73.76 \\
 & RLOO            & 0.41 & 46.15 & 64.87 & 35.33 & 40.00 & 37.67 & \underline{78.82} & 76.60 \\
 & GRPO            & 0.41 & 49.74 & 66.41 & 36.00 & \textbf{43.33} & 39.67 & 76.46 & 75.18 \\
\cmidrule(lr){2-10}
 & HarnessForge    & 0.41 & \underline{52.82} & \underline{68.10} & \textbf{42.00} & \underline{42.00} & \textbf{42.00} & \textbf{80.07} & \textbf{82.27} \\
 & \textbf{Ours} & 0.04 & \textbf{57.89} & \textbf{71.30} & \underline{40.00} & \underline{42.00} & \underline{41.00} & 76.47 & \underline{81.82} \\
\midrule
\multirow{10}{*}{Qwen3-8B}
 & ADAS            & 0 & 42.05 & 61.32 & 31.33 & 32.00 & 31.67 & 60.14 & 59.57 \\
 & AgentSquare     & 0 & 30.77 & 41.22 & 29.33 & 30.00 & 29.50 & 54.37 & 53.90 \\
 & AFlow           & 0 & 32.82 & 54.12 & 31.33 & 32.67 & 32.00 & 49.74 & 47.52 \\
 & MaAS            & 0 & 44.62 & 51.80 & 32.00 & 32.00 & 32.00 & 66.72 & 60.28 \\
 & MermaidFlow     & 0 & 47.39 & 59.31 & 32.67 & 30.67 & 31.67 & 67.46 & 62.41 \\
\cmidrule(lr){2-10}
 & SFT             & 0.27 & 48.72 & 72.20 & 40.67 & 41.33 & 41.00 & 73.80 & 73.05 \\
 & RLOO            & 0.27 & 50.77 & 72.75 & 40.00 & 39.33 & 39.67 & 73.76 & 71.63 \\
 & GRPO            & 0.27 & 51.28 & \underline{75.62} & \underline{41.33} & 42.00 & 41.67 & \underline{74.71} & \underline{73.76} \\
\cmidrule(lr){2-10}
 & HarnessForge    & 0.27 & \underline{54.87} & 74.05 & \underline{41.33} & \underline{44.33} & \underline{42.83} & \textbf{78.22} & \textbf{78.01} \\
 & \textbf{Ours} & 0.03 & \textbf{62.46} & \textbf{76.55} & \textbf{42.67} & \textbf{50.67} & \textbf{46.70} & 67.65 & 72.73 \\
\bottomrule
\end{tabular}
\end{table}

\subsection{Policy compilation on agentic benchmarks}
\label{sec:policy}

Table~\ref{tab:policy-main} reports policy compilation on the HarnessForge benchmark suite. Our single-round protocol uses the official harness pairing, evaluation pools disjoint from training, an unaligned base policy, and at least two repetitions per configuration. Under it the compiled shift alone reaches 57.89\% answer accuracy on ToolHop at 4B and 62.46\% at 8B, above the published full co-evolution numbers on the same backbones (52.82 and 54.87), and it lands within a few points of them on SearchQA and API-Bank. The dagger rows differ in protocol from the co-evolution rows above them, so we read them as context and not as a head-to-head comparison.

\begin{figure}[t]
\centering
\begin{subfigure}{0.31\linewidth}\centering
\includegraphics[width=\linewidth]{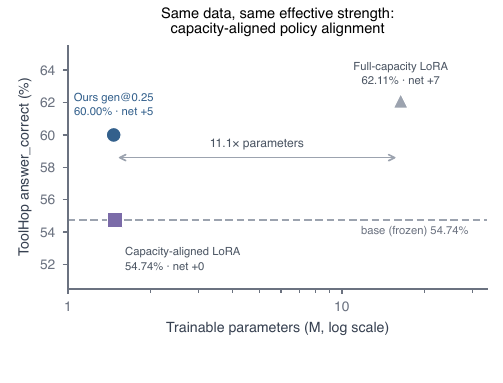}
\caption{Capacity-aligned comparison}\label{fig:capacity}
\end{subfigure}\hfill
\begin{subfigure}{0.31\linewidth}\centering
\includegraphics[width=\linewidth]{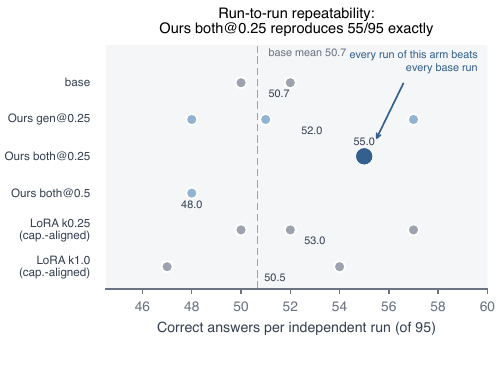}
\caption{Run-level repeatability}\label{fig:repeat}
\end{subfigure}\hfill
\begin{subfigure}{0.31\linewidth}\centering
\includegraphics[width=\linewidth]{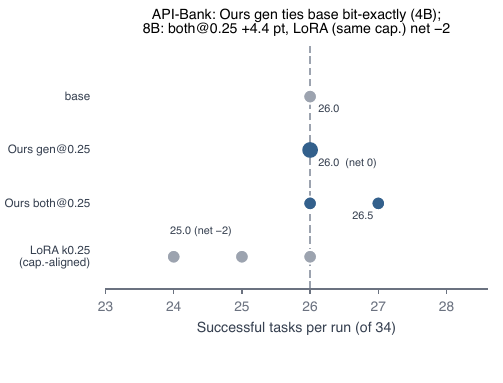}
\caption{API-Bank zero regression}\label{fig:zero}
\end{subfigure}
\caption{\textbf{Policy compilation, efficiency, stability, and safety.} (a)~Capacity alignment on ToolHop. (b)~Run-level repeatability. (c)~Zero regression on API-Bank at 4B and 8B.}
\label{fig:policy}
\end{figure}

\paragraph{Capacity-aligned comparison.}
We align three variables at once. The LoRA control acts on q/k/v with rank 3 and 1.49M parameters (1.012$\times$ ours), trains on the same 829 ToolHop decision pairs, and applies the same effective intervention multiplier (0.5). Figure~\ref{fig:capacity} shows the outcome. The shift reaches 60.00\% (paired net $+5$ over base) while the capacity-aligned LoRA nets zero, and full-capacity LoRA (rank 8, all linear projections, 16.4M parameters, 11.1$\times$ ours) reaches 62.11\%. At 8B, LoRA r11 with 3.09$\times$ our parameters matches the best shift configuration (62.63\% vs.\ 62.46\%). The efficiency gap follows from the site. A post-block parameter is read by every downstream layer, whereas an in-layer parameter is not. The shift also stays scorable at full strength ($s{=}1.0$, net $+1$), whereas merged LoRA deltas at enlarged coefficients produced no scorable outputs under the earlier protocol. The runtime strength control therefore doubles as a safety margin that preserves the tool-call contract.

\paragraph{Stability.}
Agentic evaluation under greedy decoding still varies from run to run, so single runs mislead. The \texttt{both@0.25} configuration reproduced 55/95 correct in two independent runs, an identical aggregate score despite 14 per-item flips, and it is the only configuration whose every run exceeds every base run (Figure~\ref{fig:repeat}). Its path score (0.713) is the highest of all configurations, and it triggered no rejected tool calls, which indicates that prompt-phase injection does not strain the tool-call contract. Gains also come without action inflation (14.41 steps vs.\ base 11.80 and full-capacity LoRA 14.61).

\paragraph{Zero regression where base is strong.}
On API-Bank, a short-chain and format-heavy benchmark where the base policy is already strong, the generation-phase shift ties base bit-for-bit across three base runs (0.7647), while the capacity-aligned LoRA nets $-2$ with all three runs below base (Figure~\ref{fig:zero}). At 8B every shift configuration is at or above base, and \texttt{both@0.25} adds $+4.4$ points. Preserving an already-good policy is a property that weight updates did not provide at matched capacity.

\subsection{Instruction compilation preserves the source gain}
\label{sec:behavior}

\begin{figure}[t]
\centering
\begin{subfigure}{0.478\linewidth}\centering
\includegraphics[width=\linewidth]{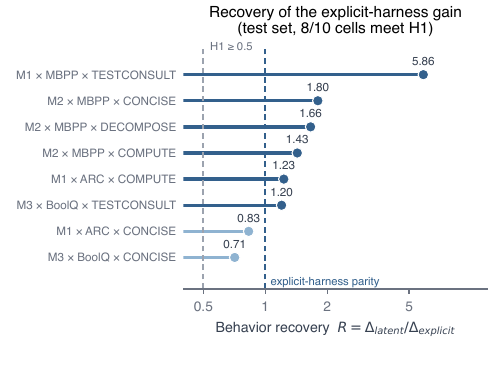}
\caption{Recovery across settings}\label{fig:recovery}
\end{subfigure}\hfill
\begin{subfigure}{0.478\linewidth}\centering
\includegraphics[width=\linewidth]{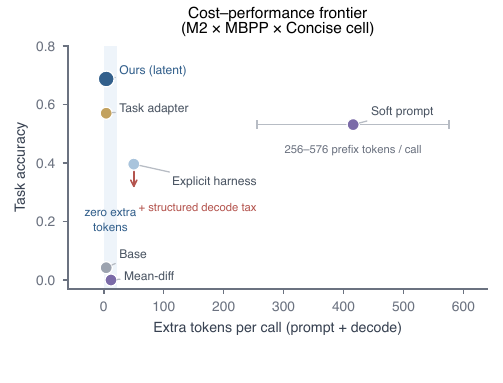}
\caption{Token cost vs.\ performance}\label{fig:frontier}
\end{subfigure}
\caption{\textbf{Instruction compilation.} (a)~Behavior recovery on the test set. (b)~Token cost against accuracy.}
\label{fig:instr}
\end{figure}

Screening first fixes which settings enter the study. Of 49 harness, model, and dataset settings, 10 show a significant explicit gain and are admitted (Appendix~\ref{app:screening}). Table~\ref{tab:seven-arm} reports the preregistered controlled test-set comparison, and Figure~\ref{fig:recovery} plots the recovery rate $R = \Delta_{\text{latent}}/\Delta_{\text{explicit}}$. Eight of the ten settings meet the preregistered criterion of $R \ge 0.5$ with the 95\% CI excluding zero.

\begin{table}[t]
\centering
\small
\setlength{\tabcolsep}{4.0pt}
\caption{Controlled test-set comparison, one touch and preregistered.
SOFT-PROMPT is parameter-matched and trained on the same distillation signal.
TASK-ADAPTER is trained on gold labels at matched capacity. $\Delta_L$ is ours
over base, and $R=\Delta_L/\Delta_E$ uses the explicit gains in
Fig.~\ref{fig:screening}. All $\Delta_L$ CIs exclude zero, and 8 of 10
enrolled settings meet the criterion $R\ge0.5$.}
\label{tab:seven-arm}
\begin{tabular}{lrrrrr}
\toprule
Setting & EXPL. & SOFT-PROMPT & TASK-ADAPTER & \textbf{OURS} & $\Delta_L$ ($R$) \\
\midrule
M1 $\cdot$ ARC $\times$ COMPUTE & 0.300 & 0.347 & 0.533 & \textbf{0.322} & $+0.082$ (1.23) \\
M1 $\cdot$ ARC $\times$ CONCISE & 0.380 & 0.376 & 0.533 & \textbf{0.362} & $+0.122$ (0.83) \\
M1 $\cdot$ MBPP $\times$ TESTCONSULT & 0.066 & 0.079 & 0.308 & \textbf{0.323} & $+0.317$ (5.86) \\
M2 $\cdot$ MBPP $\times$ DECOMPOSE & 0.410 & 0.508 & 0.562 & \textbf{0.661} & $+0.619$ (1.66) \\
M2 $\cdot$ MBPP $\times$ COMPUTE & 0.470 & 0.521 & 0.554 & \textbf{0.660} & $+0.618$ (1.43) \\
M2 $\cdot$ MBPP $\times$ CONCISE & 0.396 & 0.531 & 0.570 & \textbf{0.687} & $+0.645$ (1.80) \\
M3 $\cdot$ BoolQ $\times$ CONCISE & 0.809 & 0.833 & 0.891 & \textbf{0.774} & $+0.077$ (0.71) \\
M3 $\cdot$ BoolQ $\times$ TESTCONSULT & 0.807 & 0.822 & 0.893 & \textbf{0.826} & $+0.128$ (1.20) \\
\bottomrule
\end{tabular}
\end{table}

\noindent The clearest evidence comes from M1 on MBPP, where the compiled shift distills the test-consult behavior from only 20 correct teacher traces (teacher pass@1 is 0.073 at the format floor) and gains $+31.7$ test points. Recovery ratios divide by the explicit gain, so settings with a small explicit source produce large ratios ($R{=}5.86$ here, against an explicit gain of $+5.4$ points). We therefore read this setting by its absolute delta, and the observation we defend is that the information needed to install a behavior is far smaller than the information needed to solve the task. On M2$\times$MBPP the compiled form lands $+19$ to $+29$ points above its explicit source (0.660--0.687 vs.\ 0.396--0.470), and there the denominator is not small. The explicit harness itself gains $+0.36$ to $+0.43$ over a near-zero base. The token tax explains the difference. The explicit harness spends decode budget at every call, while the compiled behavior spends it once at compile time and nothing per call. This also places \ours{} alone at the zero-token point of the cost-performance frontier, far from parameter-matched soft prompts that pay 256--576 tokens on every call (Figure~\ref{fig:frontier}). The pattern replicates on an independent track with a base (non-instruct) 1B model, which recovers the source gain in both settings (Appendix~\ref{app:replication}).

\subsection{What the bundle carries}
\label{sec:what}

\begin{figure}[t]
\centering
\begin{subfigure}{0.31\linewidth}\centering
\includegraphics[width=\linewidth]{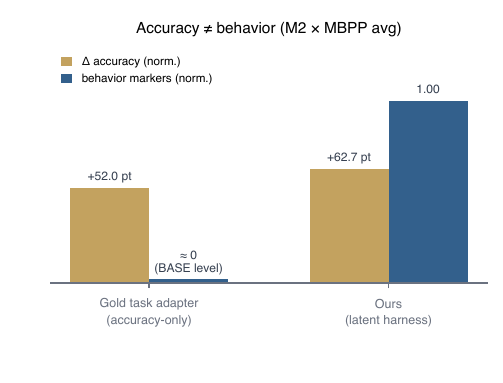}
\caption{Accuracy vs.\ behavior}\label{fig:dissoc}
\end{subfigure}\hfill
\begin{subfigure}{0.31\linewidth}\centering
\includegraphics[width=\linewidth]{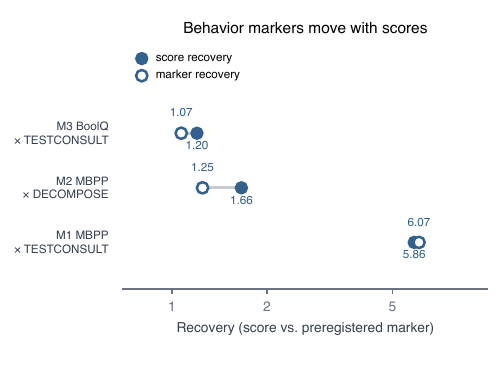}
\caption{Marker recovery}\label{fig:markers}
\end{subfigure}\hfill
\begin{subfigure}{0.31\linewidth}\centering
\includegraphics[width=\linewidth]{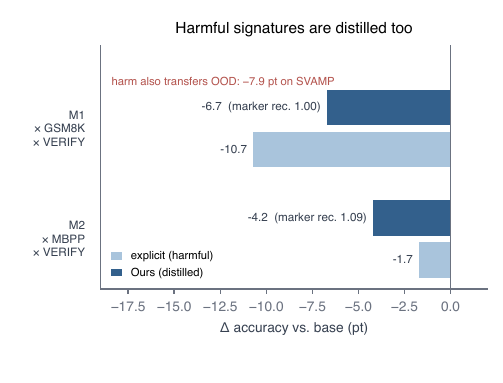}
\caption{Harmful signatures}\label{fig:harmful}
\end{subfigure}
\caption{\textbf{Three-chain attribution.} (a)~Accuracy against behavior markers (M2$\times$MBPP average). (b)~Score recovery against marker recovery. (c)~Harmful signatures compile symmetrically and transfer out of distribution.}
\label{fig:chains}
\end{figure}

Accuracy alone cannot attribute a compiled artifact. A gold-label adapter trained at matched capacity is the \emph{strongest} accuracy configuration (up to $+0.65$), yet it leaves every behavioral marker at base level (Figure~\ref{fig:dissoc}). We therefore preregistered three evidence chains. The score chain is Table~\ref{tab:seven-arm}. The marker chain shows behavior moving with scores. Decomposition steps rise from 0 to 4.6 against an explicit source of 4.5, interface alignment from 0.039 to 0.97, and answer-line rate from 0.525 to 1.00, each matching score recovery in magnitude (Figure~\ref{fig:markers}). The control chain closes the alternatives. Shuffling teacher traces across harnesses flattens every setting and every marker, and parameter-matched soft prompts trained on the same distillation signal are weaker everywhere they can be compared while paying hundreds of tokens per call.

The sharpest test of specificity is symmetric. If the bundle encodes harness semantics, harmful harnesses should compile as faithfully as helpful ones. They do. A verification harness that hurts M1 on GSM8K ($-10.7$ points explicit) distills to $-6.7$ points with verify-action marker recovery 1.00, and a harmful M2 pairing reproduces $-4.2$ with marker recovery 1.09. The negative signature even transfers out of distribution to SVAMP ($-7.9$ points, Figure~\ref{fig:harmful}). Whatever the bundle carries, it carries the harness's behavior rather than a generic improvement.

\subsection{Compiled bundles are reusable objects}
\label{sec:reusability}

\begin{figure}[t]
\centering
\begin{subfigure}{0.31\linewidth}\centering
\includegraphics[width=\linewidth]{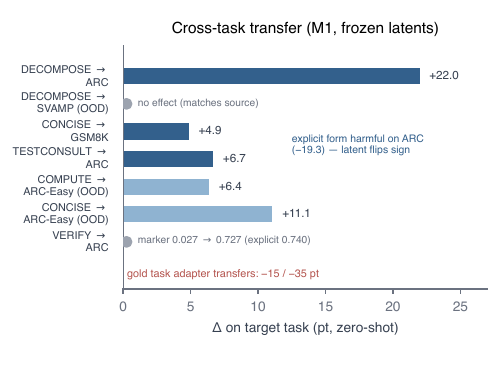}
\caption{Zero-shot transfer}\label{fig:transfer}
\end{subfigure}\hfill
\begin{subfigure}{0.31\linewidth}\centering
\includegraphics[width=\linewidth]{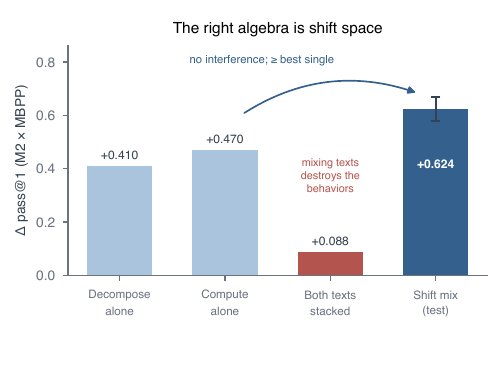}
\caption{Composition algebra}\label{fig:comp}
\end{subfigure}\hfill
\begin{subfigure}{0.31\linewidth}\centering
\includegraphics[width=\linewidth]{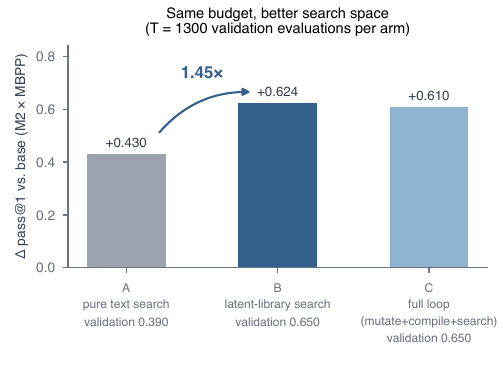}
\caption{Budget-matched search}\label{fig:search}
\end{subfigure}
\caption{\textbf{Reuse of compiled behaviors.} (a)~Cross-task transfer of frozen bundles. (b)~Text stacking against shift-space composition. (c)~Search gain at matched evaluation budgets.}
\label{fig:reuse}
\end{figure}

\paragraph{Transfer.}
Freezing five compiled bundles and evaluating seven preregistered (source $\to$ target) pairs with no retraining and no target-side tuning, five pairs transfer positively with CIs excluding zero (Figure~\ref{fig:transfer}). Decomposition transfers to ARC with $+22.0$ points, and the marker follows (steps 1.10 $\to$ 5.20). Two out-of-distribution rows into ARC-Easy recover 59\% and 74\% of their source gains. One transfer flips the sign usefully. The test-consult harness is harmful on ARC in text form ($-19.3$), yet its compiled form gains $+6.7$, because compilation distills the behavior as the teacher expresses it under that task, which differs from the instruction's literal text. The verification bundle transfers no score on ARC at all, but it moves its marker from 0.027 to 0.727 against 0.740 in explicit form, which is behavior migration on its own. Gold-label task adapters, by contrast, transfer at $-0.15$ and $-0.35$. Task skills overfit, and compiled behaviors do not.

\paragraph{Composition.}
Two compiled behaviors for M2$\times$MBPP (decompose, compute) combine exactly in shift space through the rank-16 merge. Every nonzero point of a 25-point $(\alpha,\beta)$ validation grid lands in 0.60--0.67 against a base of 0.02, and the frozen test combination reaches $+0.624$, no worse than the best single component, with both marker signatures intact. Stacking the same two instruction texts in context is destructive instead, at $+0.088$, a near-total collapse caused by mixed compliance with two mandated structures. The right algebra for behaviors is the shift space.

\paragraph{Search.}
If bundles compose, a library of them turns harness search from discrete text rewriting into continuous coefficient selection. With evaluation budgets pinned at $T{=}1300$ validation evaluations per configuration on M2$\times$MBPP, text-space search reaches $+0.430$ test gain, latent library search reaches $+0.624$, and the full loop of text mutation, compilation, and coefficient search reaches $+0.610$. The compiled library alone accounts for the 1.45$\times$ multiplier (Figure~\ref{fig:search}).

\subsection{Where the intervention acts, and when it fails}
\label{sec:mechanism}

\begin{figure}[t]
\centering
\begin{subfigure}{0.31\linewidth}\centering
\includegraphics[width=\linewidth]{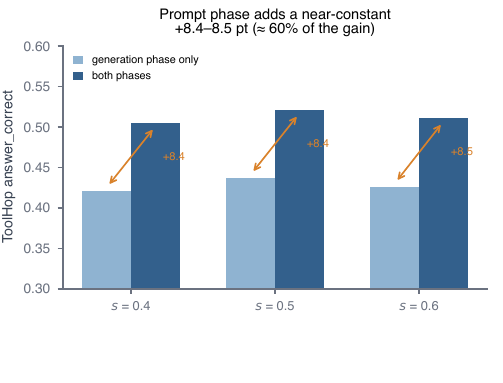}
\caption{Phase decomposition}\label{fig:phase}
\end{subfigure}\hfill
\begin{subfigure}{0.31\linewidth}\centering
\includegraphics[width=\linewidth]{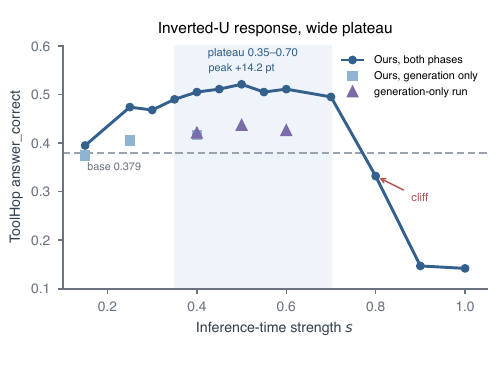}
\caption{Strength response}\label{fig:scale}
\end{subfigure}\hfill
\begin{subfigure}{0.31\linewidth}\centering
\includegraphics[width=\linewidth]{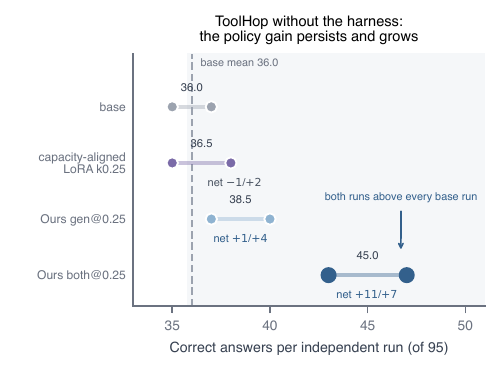}
\caption{Without the harness}\label{fig:noharness}
\end{subfigure}
\caption{\textbf{Mechanism.} (a)~Phase decomposition of the gain. (b)~Strength response with plateau and cliff. (c)~ToolHop with the harness removed, by configuration.}
\label{fig:mech}
\end{figure}

\paragraph{The gain does not depend on the harness text.}
The policy gain does not require the harness to be present. With the harness removed and the protocol otherwise unchanged, \texttt{both@0.25} still delivers $+9.5$ points on ToolHop (paired net $+11$/$+7$, with both runs above every base run), while the capacity-aligned LoRA stays at base level (Figure~\ref{fig:noharness}). Removing the harness enlarges the policy gain rather than shrinking it, which indicates partial substitution. The shift performs structural corrections that the harness would otherwise impose, and it does so for behaviors the harness text does not cover.

\paragraph{The action site is the representation layer.}
Prompt-phase injection changes every downstream computation through attention and the KV cache, while generation-phase injection changes token-level dynamics only. Decomposing the two (Figure~\ref{fig:phase}) shows that prompt positions contribute $+8.4$ to $+8.5$ points, near-constant across strengths and about 60\% of the total. Instruction compilation is the limiting case of the same fact, since the student trains on byte-identical prompts and still acquires the behavior entirely through representation-level correction.

\paragraph{Strength has a wide safe plateau and a sharp cliff.}
Sweeping 13 strength values with two repetitions each (Figure~\ref{fig:scale}) gives an inverted U with a 0.35--0.70 plateau. Within the plateau all seven points fall within 2.6 points of each other, and the peak reaches $+14.2$ over base. Beyond the plateau the failure is specific and diagnosable. Trajectories complete normally, but \texttt{has\_valid\_answer} collapses monotonically (1.000 $\to$ 0.284), a termination failure driven by prompt-side over-injection. Generation-only injection never triggers it at any tested strength. The optimal strength is condition-dependent (0.25 with the harness, 0.35--0.70 without), which is why it is exposed as a runtime control and not baked into weights.

\paragraph{What survives a scale change.}
Re-running the identical pipeline on Qwen3-8B preserves the skeleton. Zero-weight adaptation works, the parameter-efficiency ordering holds (LoRA needs 3.09$\times$ parameters to match), the failure family is the same with an earlier onset, and API-Bank remains regression-free. What moves is the gain location. ToolHop gains spread across several strengths instead of one, and SearchQA flips from no gain at 4B to $+7.5$--$9.2$ points at 8B (Appendix~\ref{app:crossmodel}). Applicability is therefore established per model-task pair, and the bundle's runtime controls are what make per-pair deployment practical.

\section{Related work}
\label{sec:related}

\paragraph{Harness engineering and self-evolution.}
Automated harness design searches over programs \citep{hu2025automated,zhang2025aflow} and over modular pipelines \citep{shang2025agentsquare,zhang2025multi}, and some variants constrain the search to safety \citep{zheng2025mermaidflow}. Self-improving systems synthesize or edit harnesses from execution feedback \citep{lou2026autoharness,shao2026harness,zhang2026self,lin2026agentic,karten2026continual}, optimize harness and model end-to-end \citep{lee2026meta}, or compile harness logic into code and skill programs \citep{ning2026code,liu2026harnessing,chen2026skvm}. Self-evolving agents typically internalize what they learn through weight updates. Examples include RL-trained reasoning from zero data \citep{huang2026r}, recursive skill-augmented RL \citep{xia2026skillrl}, memory skills \citep{zhang2026memskill}, and self-referential improvement loops \citep{yin2025godel}, while \citet{lin2026harness} disentangle harness updates from harness benefits. HarnessForge \citep{chen2026harnessforgejointharnesspolicy} co-evolves harnesses and policies and provides the benchmarks, harness pairings, and training protocol we adopt. Its training-style configurations (SFT, RLOO, GRPO) and its co-evolution objective all write gains into weights. \ours{} is complementary. It moves a fixed harness's behavioral content into activations, keeps the backbone frozen, and leaves the harness-evolution loop free to iterate on text.

\paragraph{Parameter-efficient adaptation.}
LoRA \citep{hu2021lora}, prefix-tuning \citep{li2021prefix}, and prompt tuning \citep{lester2021power} attach low-rank or prompt-side parameters to a frozen backbone. Our operator shares the low-rank parametrization but acts at the post-block junction of the residual stream. Our comparisons hold data, capacity, and effective strength fixed, so they isolate the effect of the site from that of the budget. Behavioral cloning \citep{torabi2018behavioral} and data-efficient reasoning distillation \citep{zhang2026skill} are output-mimicking relatives of ours. The differences are strict correctness filtering of the source and a preregistered behavior-level attribution protocol on the target.

\paragraph{Activation steering and representation engineering.}
Adding direction vectors to activations \citep{turner2024activation,turner2023steering}, contrastive activation addition \citep{rimsky2024steering}, representation engineering \citep{zou2023representation,bartoszcze2025representation}, inference-time intervention \citep{li2023inference}, function vectors \citep{todd2024function}, and learned representation interventions \citep{wu2024reft} study how latent-space edits change model behavior. Systematic evaluations find difference-vector methods unreliable and learned interventions stronger \citep{tan2024analysing,wu2025axbench,stoehr2024activation}, and later work extends the setting to refusal steering \citep{sheng2026alphasteer}, defense \citep{zou2024improving}, subspace models \citep{zhao2025beyond}, hallucination detection \citep{park2025steer} and decision operators \citep{li2026learningdecidereasonparameterefficient}. Our work follows the learned-intervention regime. \citet{adila2026weightupdatesactivationshifts} formalize weight updates as activation shifts. We use the post-block shift as a compilation target for external conditions and add three things that steering work has not addressed, namely behavior-level attribution, composition algebra, and transfer together with agent-benchmark policy alignment under controlled capacity.

\section{Limitations}
\label{sec:limitations}

Gain locations move with the model-task pair. ToolHop gains concentrate or spread across strengths, and SearchQA flips direction from 4B to 8B. We therefore claim per-pair applicability only, and the runtime controls exist to support per-pair tuning. Strength has a hard ceiling. Beyond the plateau the policy stops producing valid answers, the failure is prompt-side specific, and the onset moves earlier with scale, so deployments must sweep $s$ before fixing it. Our policy results are single-round alignment of a base policy to a fixed harness, not multi-round co-evolution, so the dagger rows of Table~\ref{tab:policy-main} are protocol-distinct from the published co-evolution numbers. Evaluation pools are small (95, 50, and 34 items). We mitigate this with repeated full evaluations and per-item paired analysis, but absolute deltas of a few points should be read with that limitation in mind. Instruction compilation covers short process-level instructions. Harnesses too complex to state as text are handled by the trajectory route.

\section{Conclusion}
\label{sec:conclusion}

We treated the harness as a compilation source and not as a prompt to repeat or a skill to fine-tune. A rank-8 shift at the post-block junction of a frozen backbone compiles both explicit behavior instructions and agentic-harness strategies into bundles that need no condition text, no tokens, and no weight change. Strength, phase, and mixing stay exposed at run time. Compiled behaviors recover and sometimes exceed their sources, they transfer and compose in shift space, they align policy at a fraction of LoRA's parameters under matched conditions, and they fail in a diagnosable and bounded way. The compilation view turns harness engineering from a per-call expense into a reusable asset, since every harness the community evolves becomes a candidate bundle in a shared library.

\subsubsection*{Reproducibility statement}
All numbers here come from preregistered protocols with fixed seeds, validation-only selection, single test touches, and append-only result ledgers. Policy configurations average at least two independent full evaluations. Appendix~\ref{app:protocol} details the harness registry, training configurations, the definition of each configuration, the parse protocol, and the statistical procedures. Appendix~\ref{app:crossmodel} documents cross-scale runs and protocol notes.

\subsubsection*{Ethics statement}
Compilation is semantically faithful in both directions. Harmful harnesses distill into harmful bundles as readily as helpful ones, which we report as an attribution finding without claiming a capability. Deployment should check compile sources against a registry, preregister evaluation criteria, and audit exported bundles. A practical advantage of the runtime form is that a bundle can be inspected, tuned down, or removed without touching model weights.

\bibliography{latentharness}
\bibliographystyle{iclr2027_conference}

\appendix

\section{Experimental protocol details}
\label{app:protocol}

\begin{table}[t]
\centering
\small
\caption{Models and coverage. Rank-8 post-block adapters cover all layers and all
tokens, with a SiLU bottleneck for instruction compilation and a linear map for
policy compilation. The backbone is frozen, with every parameter verified
identical to the original.}
\label{tab:setup}
\begin{tabular}{llrrl}
\toprule
Code & Model & Layers $\times$ $d$ & Trainable & Share \\
\midrule
M1 & Llama-3.2-1B-Instruct & 16 $\times$ 2048 & 524{,}288 & 0.0424\% \\
M2 & Qwen3-4B & 36 $\times$ 2560 & 1{,}474{,}560 & 0.0367\% \\
M3 & Llama-3.1-8B-Instruct & 32 $\times$ 4096 & 2{,}097{,}152 & 0.0312\% \\
v2 & Llama-3.2-1B (base) & 16 $\times$ 2048 & 524{,}288 & 0.0424\% \\
-- & Qwen3-8B (policy) & 36 $\times$ 4096 & 2{,}359{,}296 & 0.0288\% \\
\midrule
\multicolumn{5}{l}{Instruction tasks are BoolQ, ARC-Challenge, GSM8K, and MBPP.} \\
\multicolumn{5}{l}{SVAMP and ARC-Easy serve as out-of-distribution targets.} \\
\multicolumn{5}{l}{Policy benchmarks are ToolHop (n=95), SearchQA-hotpotqa (n=50), and API-Bank (n=34).} \\
\bottomrule
\end{tabular}
\end{table}

\paragraph{Harness registry.}
Seven single-pass, tool-free, process-level instructions (VERIFY, DECOMPOSE, COMPUTE, CONCISE, ELIMINATE, QUOTE, TESTCONSULT) each stay at most 60 tokens. Each is registered by a content hash with a named target error mode and preregistered markers, computed by deterministic parsing of raw outputs (verify-action rate, decomposition-step count, arithmetic-expression rate, answer-line rate, option-coverage rate, evidence-quote rate, interface-alignment rate, clean-stop rate). By construction, markers never fire on base-prompt echo, and they are checked to that effect.

\paragraph{Instruction compilation.}
Teacher traces are greedy generations of the frozen model with the harness prepended. All outputs are cached, and correctness filtering selects the imitation target at training time. The student trains on $n{=}\min(600, |\text{pool}|)$ examples for 2 epochs at lr $9{\times}10^{-4}$ with 3 seeds $\{42,43,44\}$. The loss is prompt-masked token cross-entropy on filtered traces for generation tasks or an option-level KL against cached teacher distributions for multiple choice, with option-string mirroring locked by unit test. Hard per-step assertions cover three properties. Backbone parameters stay identical to the original, the optimizer holds adapter parameters only, and no harness substring appears in any student input while harness text is present in every teacher prompt.

\paragraph{Policy compilation.}
Strict-tier decision pairs come from harness-conditioned rollouts (ToolHop 1{,}042 pairs, 829 train / 213 val, SearchQA 670, API-Bank 327). Training uses prefix-masked cross-entropy on completions with lr $10^{-4}$, 3 epochs, batch size 1 with gradient accumulation 2, max length 4{,}096, and bf16 with gradient checkpointing. The LoRA comparison configurations use identical hyperparameters. The capacity-aligned configuration is rank 3 on q/k/v (1.49M parameters, 1.012$\times$) with effective multiplier $\tfrac{\alpha_L}{r_L}k = 0.5$ matched to the shift's $s\,\alpha = 0.5$. The full-capacity configuration is rank 8 on all linear projections (16.4M), and at 8B the reference configuration is rank 11 (7.30M, 3.09$\times$).

\paragraph{Evaluation.}
Decoding is greedy throughout. Instruction settings use dataset-native extract-and-match or sandboxed pass@1 with preregistered extraction, and we report paired bootstrap 95\% CIs with $B{=}10{,}000$. Policy runs use the HarnessForge evaluator fields as produced, with at least 2 independent full evaluations per adopted configuration, per-item paired net differences, and a bounded deterministic parse protocol that repairs once and then rejects, identically across configurations. Rejection counts are reported as behavioral signals. Injection validity is verified per configuration by the scale-signature check, in which bit-identical metrics across strengths indicate silent non-injection.

\paragraph{Configuration definitions (instruction track).}
BASE applies no intervention. EXPLICIT prepends the harness text and serves as the recovery denominator. SOFT-PROMPT is a parameter-matched soft prefix trained on the same distillation signal. TASK-ADAPTER uses gold-label cross-entropy at matched capacity. OURS distills without gold labels. SHUFFLED pairs teacher traces with the wrong harness and serves as the semantic-ablation control.

\section{Explicit-harness screening and the MBPP format floor}
\label{app:screening}

\begin{figure}[h]
\centering
\includegraphics[width=0.49\linewidth]{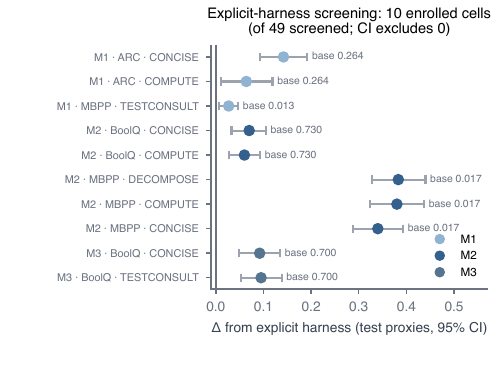}\hfill
\includegraphics[width=0.49\linewidth]{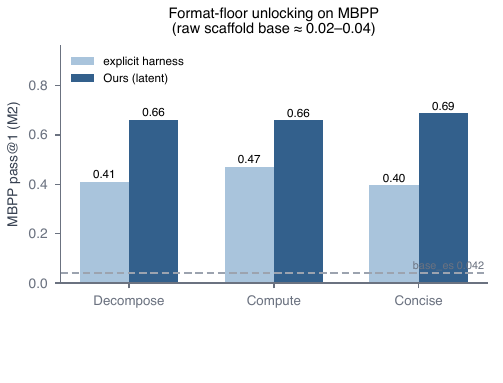}
\caption{Screening forest (left) and the MBPP format-floor unlock (right).}
\label{fig:screening}
\end{figure}

\noindent Screening evaluates 49 harness, model, and dataset settings with three validation draws each and admits the 10 settings whose explicit gain is positive with a 95\% CI excluding zero (Figure~\ref{fig:screening}, left, where each interval is one admitted setting). Settings whose explicit form does not help are recorded as not informative and never enter a recovery claim, by preregistration. The recovery rate is therefore never computed against a source that produces no gain. The admitted settings concentrate where the base model has a diagnosable failure mode. Task difficulty alone does not predict admission. On MBPP all three models sit at a format floor, with base pass@1 near 0.01--0.04, since the 3-shot scaffold almost never elicits extractable code. Structural harnesses lift M2 to 0.34--0.42 while the corresponding markers move by design, with decomposition steps going 0 $\to$ 4.2 and interface alignment 0.057 $\to$ 0.646 (Figure~\ref{fig:screening}, right). M3 passes a compliance pre-check with all three predicted markers moving in their predicted directions (evidence quotes 0.285 $\to$ 0.755, steps 0.09 $\to$ 2.03, verify actions 0.035 $\to$ 0.980), which is what makes the distillation targets of Section~\ref{sec:behavior} unambiguous. Reading the forest by model also previews the recovery pattern. Explicit gains are largest where the base is weakest, so the largest recoveries in Table~\ref{tab:seven-arm} come from the hardest starting points and not from easy settings.

\begin{figure}[h]
\centering
\includegraphics[width=0.49\linewidth]{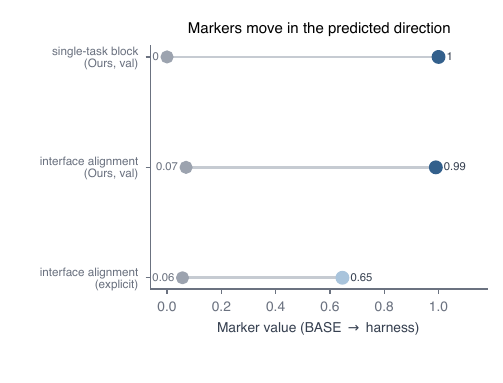}\hfill
\includegraphics[width=0.49\linewidth]{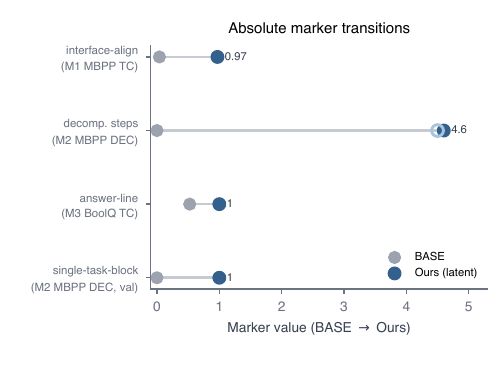}
\caption{MBPP marker displacements under explicit harnesses (left) and absolute marker transitions from base to the compiled form (right).}
\label{fig:appmbpp}
\end{figure}

\noindent Figure~\ref{fig:appmbpp} shows the two marker views that complement the score tables. The left panel tracks marker displacement under explicit harnesses in the MBPP family and confirms that the structural instructions move behavior first (steps, interface alignment, single-task blocks) while scores follow. The right panel shows the same quantities for the compiled form against base, with the explicit source as a reference point. The compiled markers reach or exceed the explicit references (steps 4.6 vs.\ 4.5, interface alignment 0.97, single-task blocks 1.00), which is the marker-level statement of the recovery result. The behavior is present, and the advantage is more than a score difference.

\section{Attribution details and per-model discussion}
\label{app:attribution}

Table~\ref{tab:seven-arm} compresses three model-specific patterns. On M1, the two ARC settings recover the explicit gain (ratios 0.83 and 1.23 on moderate denominators), while the MBPP setting adds $+31.7$ points from only 20 correct teacher traces. Its ratio of 5.86 reflects the small $+5.4$-point explicit denominator, so we read the setting by its absolute delta. The parameter-matched soft-prompt control trained on the same signal reaches 0.079 and pays 512 tokens per call, which shows that the distillation signal is necessary and that the activation site makes it sufficient. On M2, all three MBPP settings exceed their explicit sources (0.660--0.687 vs.\ 0.396--0.470). The explicit harness spends decode budget on mandated structure at every call within a 512-token limit, while the compiled form spends that budget on code. On M3, the two BoolQ settings bracket the criterion. Concise answering recovers 0.71 of a large explicit gain (0.809 $\to$ 0.891 explicit, ours 0.774), while test-consult recovers 1.20 with 98\% validation recovery and a seed range of 0.015, the most stable setting in the study. Across all settings the TASK-ADAPTER control is the strongest accuracy configuration and the weakest behavioral one, which is the dissociation that motivates the three-chain protocol in Section~\ref{sec:what}.

\section{Recovery summaries and independent replication}
\label{app:replication}

\begin{figure}[h]
\centering
\includegraphics[width=0.49\linewidth]{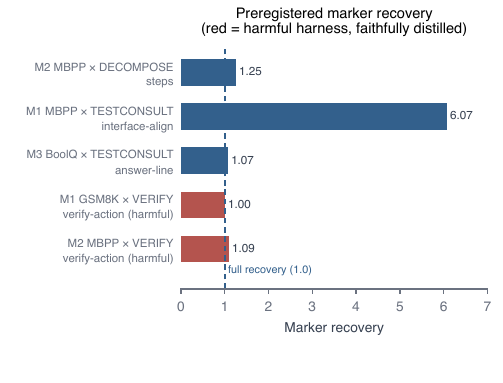}\hfill
\includegraphics[width=0.49\linewidth]{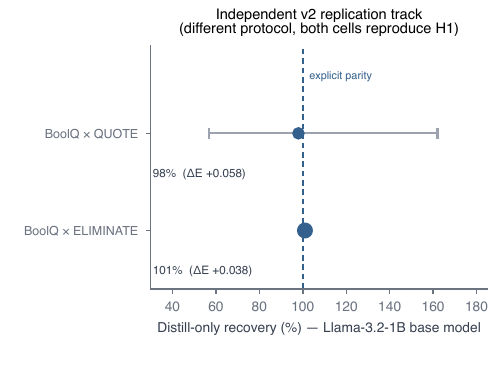}
\caption{(a)~Preregistered marker recoveries across settings. (b)~Independent replication on the base 1B model.}
\label{fig:appmarkers}
\end{figure}

\noindent Figure~\ref{fig:appmarkers} (left) collects marker recoveries across settings. The highlighted entries are the two harmful harnesses, whose verify-action behavior is reproduced at recovery 1.00 and 1.09 while their scores go negative. A generic task adaptation cannot produce a faithfully negative behavioral signature, so we read this panel as the most discriminating attribution display in the study. The right panel reports the replication track, which re-runs the full pipeline on Llama-3.2-1B (base, non-instruct) under an independent protocol with the complete BoolQ test set ($n{=}654$) and three training seeds. Both enrolled settings replicate, QUOTE at 98\% recovery (CI $[+57, 162]\%$) and ELIMINATE at 101\%, so the effect does not depend on instruct tuning or on one model family.

\section{Harmful-signature transfer and the task-skill contrast}
\label{app:harmful}

\begin{figure}[h]
\centering
\includegraphics[width=0.49\linewidth]{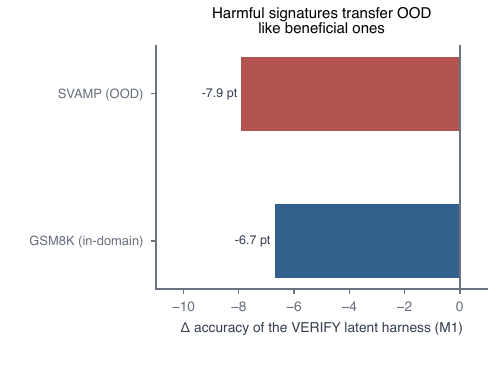}\hfill
\includegraphics[width=0.49\linewidth]{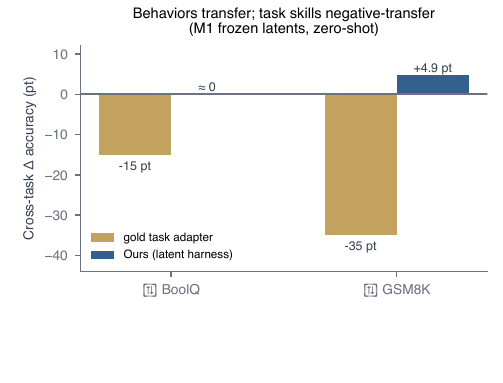}
\caption{(a)~Harmful-signature transfer to SVAMP. (b)~Transfer contrast between compiled behaviors and gold-label task adapters.}
\label{fig:apprep}
\end{figure}

\noindent Two observations close the attribution argument. The harmful VERIFY signature is mobile. Distilled from a GSM8K pairing at $-6.7$ points, it transfers out of distribution to SVAMP at $-7.9$, the same mobility positive signatures show. Harm and help moving through the same channel is strong evidence that the bundle encodes harness semantics rather than task utility. The transfer profiles also differ in kind (Figure~\ref{fig:apprep}, right). Task adapters trained on gold labels transfer negatively ($-0.15$ to BoolQ, $-0.35$ to GSM8K), while compiled behaviors transfer at zero or positive gain. Bundles behave like reusable behaviors, and task adapters behave like overfit task skills.

\section{Composition and search details}
\label{app:composition}

\begin{figure}[h]
\centering
\includegraphics[width=0.49\linewidth]{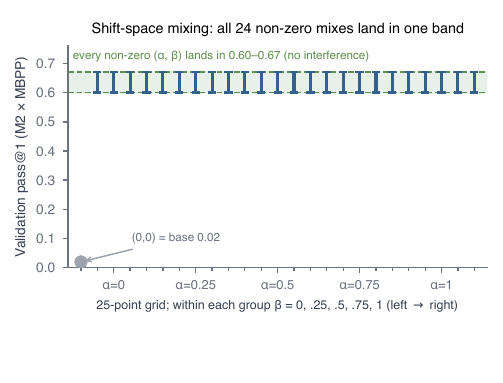}\hfill
\includegraphics[width=0.49\linewidth]{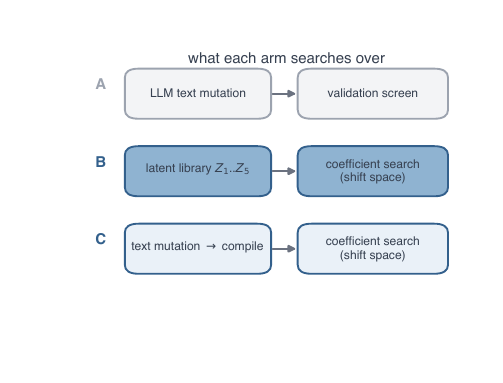}
\caption{(a)~The 25-point shift-space composition grid. (b)~The three budget-matched search configurations.}
\label{fig:appcomp}
\end{figure}

\noindent The composition grid (Figure~\ref{fig:appcomp}, left) sweeps $\alpha,\beta \in \{0, 0.25, 0.5, 0.75, 1\}$ on validation. Every nonzero point lands in 0.60--0.67 against a base of 0.02, so no mixing ratio interferes with either component. The merge that realizes $\alpha Z_a + \beta Z_b$ is exact, using rank-16 vertical and horizontal stacking with the coefficients on the up blocks and a tensor-level error of at most $10^{-5}$. Composing in parameter space after orthogonal alignment instead deviates from the shift-space composition by roughly 37\% relative error and loses over half the marker signal, which is why we locate the correct algebra in shift space. The search configurations (right) are text-only mutation and screening over instruction rewrites, coefficient search over a fixed compiled library with nonnegative coefficients capped at 2, and the full loop that mutates text, compiles it, then searches coefficients. All three receive $T{=}1300$ validation evaluations. The library configuration wins because composition makes its search space continuous and reusable instead of discrete and disposable.

\section{Policy-compilation details}
\label{app:policydetail}

\begin{figure}[h]
\centering
\includegraphics[width=0.325\linewidth]{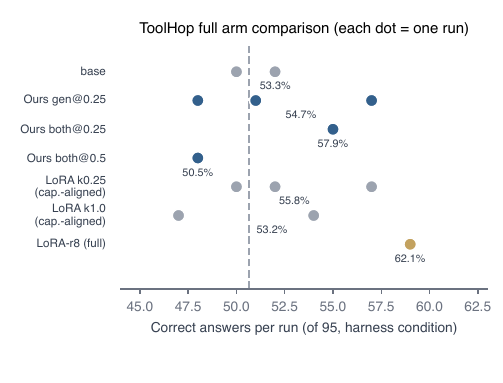}\hfill
\includegraphics[width=0.325\linewidth]{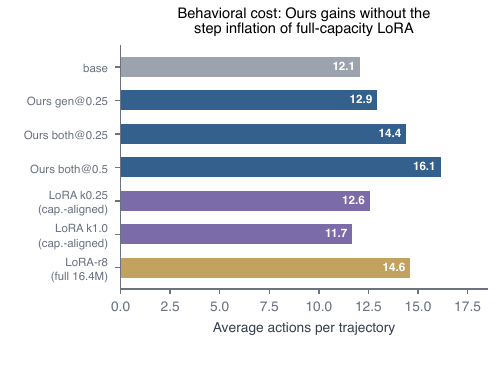}\hfill
\includegraphics[width=0.325\linewidth]{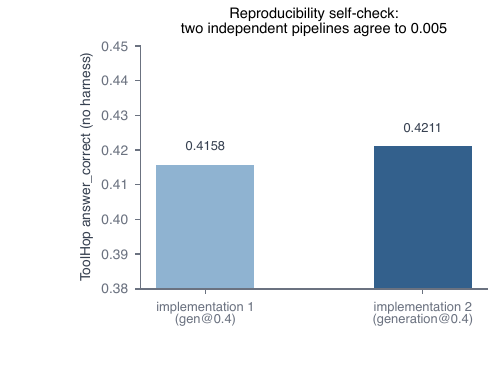}
\caption{(a)~All ToolHop configurations, with runs as points. (b)~Action counts per configuration. (c)~Reproducibility self-check.}
\label{fig:apptoolhop}
\end{figure}

\noindent Figure~\ref{fig:apptoolhop} (left) shows every ToolHop configuration including the full-capacity LoRA reference. Each point is one full evaluation, and the spread of single runs, up to $\pm5$--9 items, is why we average repeats and report paired net differences instead of single numbers. The middle panel shows what the gains cost behaviorally. The shift's \texttt{both@0.25} gains come with 14.41 average actions against base 11.80, while full-capacity LoRA inflates to 14.61 at over ten times the capacity. The generation-phase shift at the same strength adds only 7\%, so the gains do not come from longer trajectories. The right panel is the reproducibility self-check, where two independent implementations of the same configuration differ by 0.005 in the mean. That bounds run-to-run noise near half a point on this task, an order of magnitude below the effects we interpret.

\section{Strength response and failure staging}
\label{app:scale}

\begin{figure}[h]
\centering
\includegraphics[width=0.49\linewidth]{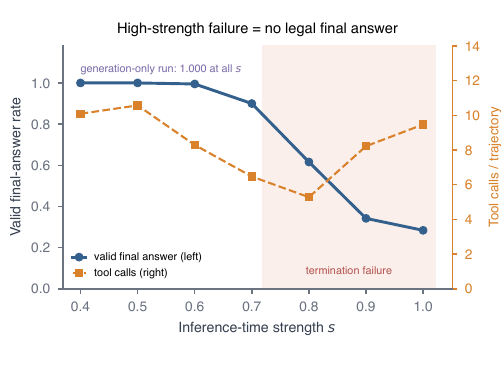}\hfill
\includegraphics[width=0.49\linewidth]{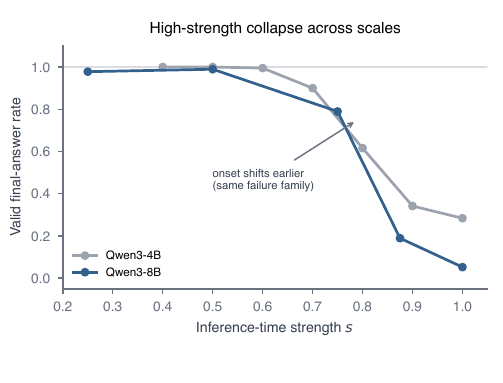}
\caption{(a)~Failure staging at high strength. (b)~The same failure family at 4B and 8B.}
\label{fig:appfail}
\end{figure}

\noindent The high-strength failure proceeds in two stages (Figure~\ref{fig:appfail}, left). Between $s{=}0.7$ and $0.8$ the policy terminates early, with action counts falling from 10.6 to 6.5 to 5.3 while \texttt{has\_valid\_answer} begins to slip. From $s{=}0.9$ the behavior degenerates into cycling, where action counts rebound to 8--9.5 but valid answers fall to 0.34 and then 0.28. Trajectories always complete and tool-call parsing never fails, so this is a policy-level termination failure. It is also prompt-side specific. Generation-only injection holds \texttt{has\_valid\_answer} at 1.000 at every strength tested. The 8B curve (right) shows the same family with an earlier onset (collapse by $s{=}0.875$), consistent with the interpretation that the phase boundary is set by the ratio of shift magnitude to the policy's behavioral margin, which shrinks as models get stronger.

\section{Cross-scale account}
\label{app:crossmodel}

\begin{table}[t]
\centering
\small
\setlength{\tabcolsep}{3.0pt}
\caption{Cross-scale account for Qwen3-4B against Qwen3-8B. The method skeleton is
invariant. Gain locations (task $\times$ phase $\times$ scale) move with the
model $\times$ task pair.}
\label{tab:crossmodel}
\begin{tabular}{@{}p{2.0cm}p{3.9cm}p{3.9cm}p{3.0cm}@{}}
\toprule
Dimension & 4B & 8B & Verdict \\
\midrule
ToolHop \texttt{both} gain & both@0.25 reproduces 55/95 ($+4.3$) & multi-band: both@0.5 $+4.6$ pt, both@0.25 $+4.0$ pt & gain location moves with scale \\
SearchQA gain & none (Hotpot subset) & Overall $+7.5$--$9.2$ pt (Hotpot $+11$--$15$) & direction flips with scale \\
Phase conclusion & \texttt{both} carries the stable gain ($\approx$60\% prompt-side) & \texttt{gen} $\approx$ \texttt{both} on ToolHop and \texttt{both} wins on SearchQA & task $\times$ scale dependent \\
LoRA comparison & same-capacity LoRA $\leq$ base everywhere & LoRA r11 (3.09$\times$ params) matches Ours & Ours matches LoRA with $\leq$1/3 params \\
Scale safe interval & plateau 0.35--0.70 and cliff $\geq 0.8$ & narrows to 0.25--0.5 and collapse $\geq 0.875$ & same failure family, earlier onset \\
API-Bank & zero regression (bit-exact tie) & all configurations at or above base (both@0.25 $+4.4$ pt) & zero regression across scales \\
\bottomrule
\end{tabular}
\end{table}

\begin{figure}[h]
\centering
\includegraphics[width=0.49\linewidth]{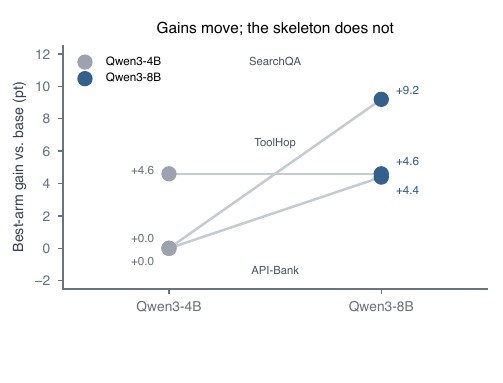}\hfill
\includegraphics[width=0.49\linewidth]{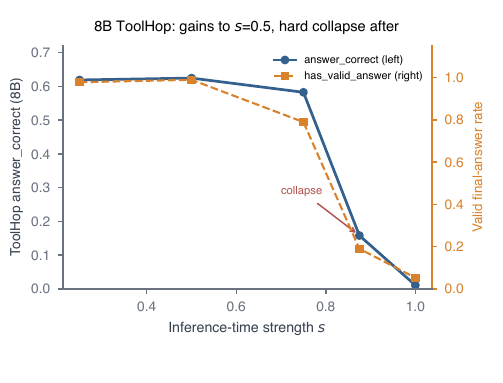}
\caption{(a)~Gain locations across benchmarks and scales. (b)~8B ToolHop across strengths.}
\label{fig:appcross}
\end{figure}

\noindent The cross-scale protocol moves the same data pipeline, hyperparameters, and evaluation protocol to Qwen3-8B unchanged, in two rounds (our 8B bundle, then the official 8B harness package). The official aligned policy weights were unreleased, so both rounds are harness $\times$ base-policy controlled comparisons. On ToolHop the gain survives at both scales, $+4.0$ to $+4.6$ points with the official pairing giving $+6.3$, but it spreads across several strengths instead of concentrating in one configuration. On SearchQA the direction flips, with no gain at 4B and $+7.5$--$9.2$ points overall at 8B carried by the Hotpot subset ($+11$--$15$). On API-Bank the zero-regression property strengthens with scale, and every configuration is at or above base, with \texttt{both@0.25} at $+4.4$. The LoRA comparisons stay directionally consistent. At 8B, LoRA r11 with 3.09$\times$ parameters reaches the same band (62.63\% vs.\ 62.46\% on ToolHop, an identical tie at 0.415 on official-protocol SearchQA) and never falls below at equal capacity. What moves across scales is the gain location (task $\times$ phase $\times$ strength). What holds is the skeleton, which covers zero-weight feasibility, the parameter-efficiency ordering, the failure family, and zero regression.

\begin{figure}[h]
\centering
\includegraphics[width=0.49\linewidth]{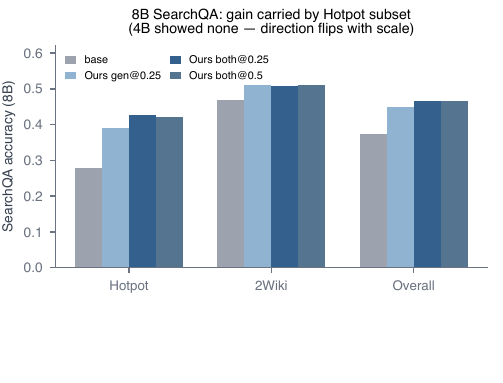}\hfill
\includegraphics[width=0.49\linewidth]{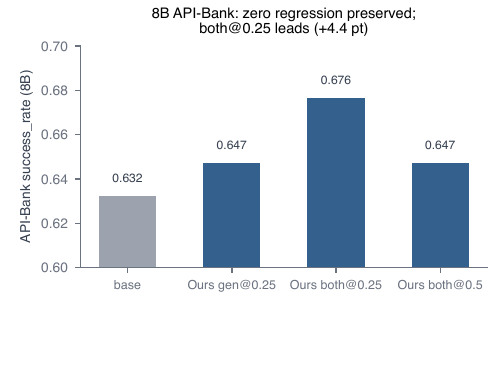}
\caption{(a)~8B SearchQA by subset. (b)~8B API-Bank by configuration.}
\label{fig:app8b}
\end{figure}

\noindent Figure~\ref{fig:app8b} splits the two remaining 8B results. SearchQA gains concentrate on Hotpot while 2Wiki moves 2--4 points, within run noise. The official-package round reproduces the Hotpot gain and an overall tie with LoRA r11, so the SearchQA conclusion we draw is subset-specific and scale-dependent. API-Bank at 8B shows the same ordering as at 4B with the signs improved. \texttt{api\_name\_accuracy} is identical across configurations (0.9412), which indicates that the injection does not disturb format selection, and the accuracy edge comes from argument accuracy.

\section{LLM usage disclosure}
\label{app:llm}
Large-language-model tools were used for editing and formatting support during manuscript preparation. All research questions, experimental design, runs, analysis, and claims are the authors' own, and all reported numbers were produced by the experimental pipeline described above.

\end{document}